\documentclass{article} 
\usepackage{colm2024_conference}

\usepackage{microtype}
\usepackage{graphicx}
\usepackage{algorithmic}
\usepackage{algorithm}
\usepackage{hyperref}
\usepackage{url}
\usepackage{booktabs}
\usepackage[labelfont=bf,font=small]{caption}

\usepackage{amsmath,amsfonts,bm}

\newcommand{\boldW}{\mathbf{W}}
\newcommand{\boldX}{\mathbf{X}}
\newcommand{\boldY}{\mathbf{Y}}
\newcommand{\boldZ}{\mathbf{Z}}

\newcommand{\boldz}{\mathbf{z}}

\newcommand{\reals}{\ensuremath{\mathbb{R}}}

\DeclareMathOperator*{\round}{round}

\def\eqref#1{equation~\ref{#1}}

\def\1{\bm{1}}

\DeclareMathAlphabet{\mathsfit}{\encodingdefault}{\sfdefault}{m}{sl}
\SetMathAlphabet{\mathsfit}{bold}{\encodingdefault}{\sfdefault}{bx}{n}

\DeclareMathOperator{\clamp}{clamp}
\let\round\relax
\DeclareMathOperator{\round}{round}
\DeclareMathOperator{\diag}{diag}

\definecolor{darkblue}{rgb}{0, 0, 0.5}
\hypersetup{colorlinks=true, citecolor=darkblue, linkcolor=darkblue, urlcolor=darkblue}

\usepackage{cleveref}

\title{Mitigating the Impact of Outlier Channels for Language\\ Model 
Quantization with Activation Regularization}

\colmfinalcopy 
\begin{document}

\maketitle
\vspace{-2cm}
\begin{center}

\textbf{Aniruddha~Nrusimha}$^1$\quad
\textbf{Mayank~Mishra}$^2$\quad
\textbf{Naigang~Wang}$^3$\quad
\\
\textbf{Dan~Alistarh}$^{4,5}$\quad
\textbf{Rameswar~Panda}$^2$\quad
\textbf{Yoon~Kim}$^1$\quad

$^{1}$Massachusetts Institute of Technology\quad
$^{2}$MIT-IBM~Watson~AI~Lab\quad
\\
$^{3}$IBM~Research\quad
$^{4}$IST~Austria\quad
$^{5}$Neural Magic, Inc.\quad

\texttt{anin@mit.edu}
\end{center}

\begin{abstract}
\vspace{-2mm}
We consider the problem of accurate quantization for language models, where both the weights and activations are quantized to 4 bits per parameter with uniform quantization, the lowest bitwidth format natively supported by existing GPU hardware. In this context, the key challenge is \emph{activation quantization}: it is known that language models contain outlier channels whose values on average are orders of magnitude higher than than other channels, which prevents accurate low-bitwidth quantization with known techniques. We systematically study this phenomena  and find that these outlier channels emerge early in training, and that they occur more frequently in layers with residual streams. We then propose a simple strategy which regularizes a layer's \emph{inputs} via quantization-aware training (QAT) and its \emph{outputs} via activation kurtosis regularization. We show that regularizing both the inputs and outputs is crucial for preventing a model's ``migrating'' the difficulty in input quantization to the weights, which makes post-training quantization (PTQ) of weights more difficult. When combined with weight PTQ, we show that our approach can obtain a W4A4 model with integer quantization that performs competitively to the standard-precision W16A16 baseline.\footnote{Code is available at \url{https://github.com/aninrusimha/qat-pretrain}}
\end{abstract}

\vspace{-5mm}
\section{Introduction} 
\vspace{-2mm}

Large language models (LLM) have been shown to contain \emph{outlier channels}, i.e., feature dimensions whose values are orders of magnitude higher than the others. These outlier channels are known to be highly correlated with strong model performance \citep{kovaleva2021bert,puccetti-etal-2022-outlier}, but  pose significant challenges from a model compression perspective, for instance via post-training quantization (PTQ)~\citep{dettmers2022llmint8,xiao2023smoothquant,Wei:NeurIPS22}. Concretely, to enable the use of low-bitwidth integer matrix multiplications---which  can lead to significant speed-ups---{both} the activations and the weights need to be quantized. However the presence of  high outlier values in the model activations results in high quantization errors, and thus overall poor PTQ accuracy (see, e.g.,  \citet{xiao2023smoothquant}).

To mitigate the effect of outlier channels  for activation quantization at the per-tensor level, existing works have explored various approaches, including keeping some of the computations in higher precision \citep{dettmers2022llmint8,ashkboos2023towards,zhao2023atom}, or  ``migrating'' the difficulty of quantizing outlier channels to other parts of the model \citep{xiao2023smoothquant,wei2023outlier,liu2023qllm}. While the above strategies have been effective for achieving INT8 activation quantization, INT4 quantization with PTQ methods remains an open challenge, with current methods still facing nontrivial degradations in perplexity \citep{wu2023understanding,shao2023omniquant,yuan2023rptq}.

\begin{figure*}[t]
\vspace{-14mm}
    \centering
    \includegraphics[width=.85\textwidth]{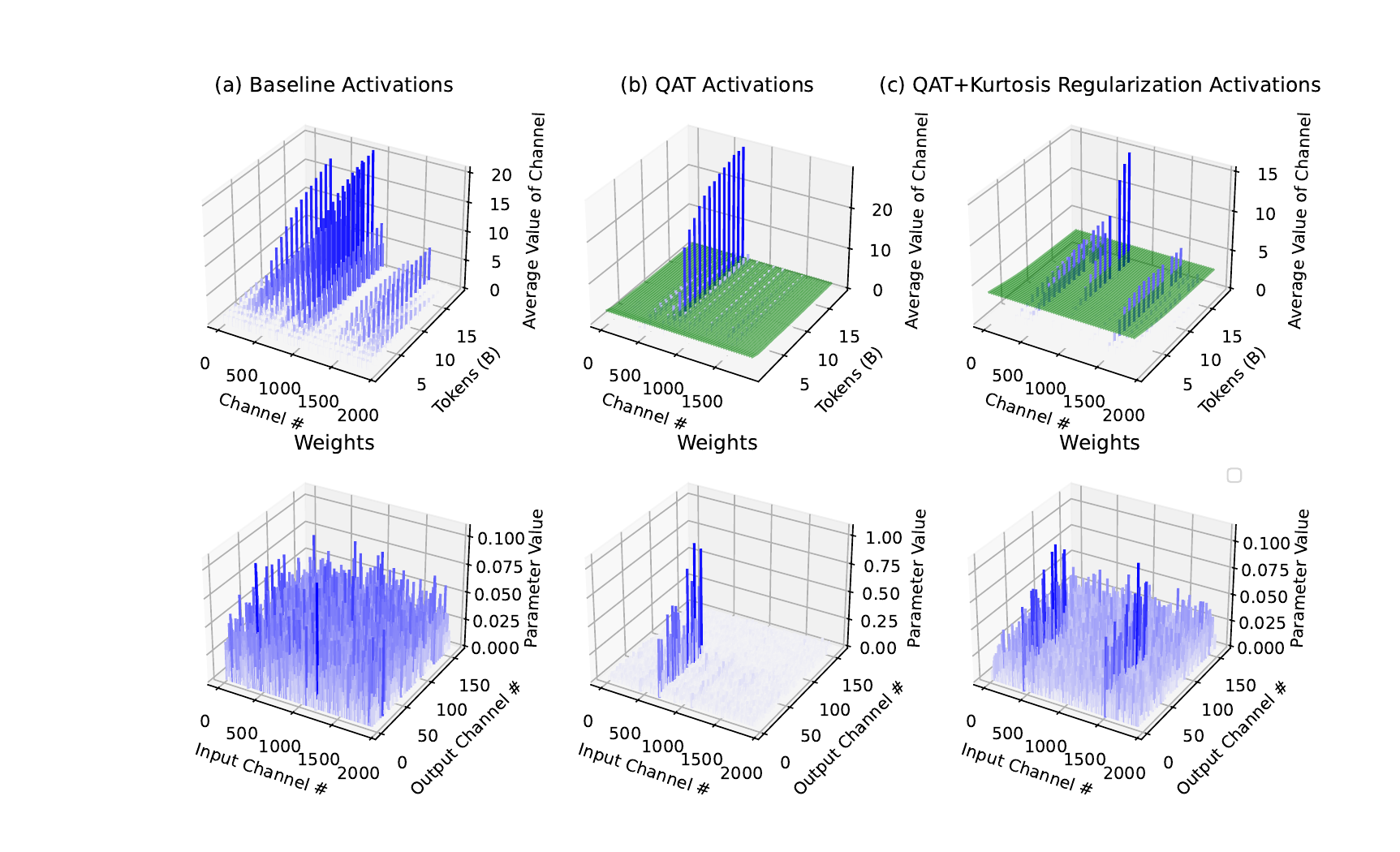}
    \vspace{-5mm}
    \caption{(Top) Average of the absolute activation values of a KV projection layer for a 1B  language model trained with (a) standard training, (b) QAT with learned clipping values in the input layer, and  (c) QAT on the inputs and kurtosis regularization on the layer's outputs.  For the QAT runs, we show the learned clip value as a green 2d manifold. (Bottom) Parameter values of individual weights in the KV projection of the same layer corresponding to each model after training.  QAT-only training results in the model's weights' becoming harder to quantize, whereas kurtosis  regularization mitigates this.}
    \vspace{-4mm}
    \label{fig:one}
\end{figure*}

In this work, we perform an empirical study of outlier channel phenomena from a pretraining perspective. We find that dimensions with outlier channels emerge relatively early in training (see \cref{fig:one}(a), top), suggesting that their mitigation requires early intervention. These outlier channels are particularly prevalent in the  output projection layer of the first layer, as well as the query-key-value projection layers of the other layers.  Next, we explore a simple  strategy that regularizes a layer's input and output. On the input side, we show that a quantization-aware training (QAT) approach which \emph{learns} the  clipping values for each activation layer \citep{choi2018pact,bhalgat2020lsq} is effective at controlling the number of outlier channels, in addition to mitigating the effect of outliers through clipping (see \cref{fig:one}(b), top).
However, while this approach can train a W16A4 model that has similar perplexity to a W16A16 model, post-training weight quantization to W4A4 results in nontrivial perplexity degradations, due to the  model's \emph{weights} now becoming more difficult to quantize (see \cref{fig:one}(b), bottom). We thus additionally regularize the kurtosis of a layer's \emph{output}, which discourages the creation of outliers wholesale. Specifically, this discourages the layer's \emph{weights} having   pathologically large rows (\cref{fig:one}(c), bottom). 

Putting all these elements together,  we show that we can train a  language model at moderate scale (1 billion parameter models trained on 20 billion tokens) whose W4A4 perplexity is competitive  to the  standard-precision W16A16 baseline.

\vspace{-3mm}
\section{Background and Related Work}
\vspace{-2mm}
\subsection{Uniform Quantization \& Quantized Matmuls}
\vspace{-2mm}
\label{background:quantization}
We focus on uniform quantization, where the quantized values are evenly spaced between an interval range.
Formally, for a given matrix $\mathbf{A} \in \reals^{n \times n}$ that we wish to quantize to $b$ bits, let $c^-$ and $c^+$ be the pre-defined (or learned) clipping values. The quantization function $Q: \reals^{n\times m} \to \mathbb{Z}^{n \times m}$ is then given by, \vspace{-2mm}
\begin{align*}
    Q({\mathbf{A}}) =   \round( s \times    \clamp( \mathbf{A}, c^-, c^+) + z ),
\end{align*}
where $s = \frac{2^b-1}{c^+ - c^-}$ is the scale factor and $z = \round(s \times c^{-})$ is the (optional) zero-point offset.
This function, which can be generalized to different granularities of $\mathbf{A}$ (e.g., rows, columns or subgroups) transforms the entries of $\mathbf{A}$ into integers between $[0, 2^b-1]$. 

The quantized matrix $\mathbf{Q}_\mathbf{A} = Q(\mathbf{A})$ can be utilized in two different ways.
First, the value can be dequantized to its original precision via $\widehat{\mathbf{A}} = \frac{1}{s}(\mathbf{Q}_\mathbf{A}-z)$ before multiplication. This method is typically used by pure weight quantization schemes, which multiply in the precision the model was trained in.  Weight-only quantization can reduce a model's memory footprint, and insofar as LLM inference  is often memory bound, it  can  also enable  faster inference by reducing the amount of time spent on  memory operations during the forward pass \citep{lin2023awq,frantar2024marlin}. However, the fact that the actual matmul is done in high precision is a fundamental limitation of weight-only quantization.

Second, the quantized values can be directly used for the matrix multiplication.
Let $\mathbf{Q}_{\mathbf{U}} = Q(\mathbf{U})$, $\mathbf{Q}_{\mathbf{V}} = Q(\mathbf{V})$ be the quantized versions of $\mathbf{U} \in \reals^{n \times k}$, $\mathbf{V} \in \reals^{k \times m}$ with the respective scaling factors $s_\mathbf{U}, s_{\mathbf{V}}$ and offsets $z_{\mathbf{U}}, z_{\mathbf{V}}$.
We can  approximate $\mathbf{U}\mathbf{V}$ with
\begin{align*}
   \label{eq:intmatmul} 
    \mathbf{U}\mathbf{V} &\approx  \widehat{\mathbf{U}}\widehat{\mathbf{V}} = \frac{1}{s_\mathbf{U} s_\mathbf{V}} \times (\mathbf{Q}_{\mathbf{U}} - z_\mathbf{U})  ( \mathbf{Q}_{\mathbf{V}} - z_\mathbf{V}),
\end{align*}
where we can make use of low-precision matmuls for $(\mathbf{Q}_{\mathbf{U}} - z_{\mathbf{U}})(\mathbf{Q}_{\mathbf{V}} - z_\mathbf{V})$.
In cases where the \emph{rows} of $\mathbf{U}$ and \emph{columns} of $\mathbf{V}$ are quantized separately  with the corresponding scaling vectors $\mathbf{s}_{\mathbf{U}} \in \reals^{n}, \mathbf{s}_{\mathbf{V}} \in \reals^m$ and offset vectors $\mathbf{z}_{\mathbf{U}} \in \mathbb{Z}^{n},\mathbf{z}_{\mathbf{V}} \in \mathbb{Z}^m$, we can still make use of integer matmuls since  $\widehat{\mathbf{U}}\widehat{\mathbf{V}}$    is given by 
\begin{align*}
\diag(\mathbf{s}_\mathbf{U})^{-1} ( \mathbf{Q}_{\mathbf{U}} - \mathbf{z}_\mathbf{U} \otimes \mathbf{1}_k)( \mathbf{Q}_{\mathbf{V}} - \mathbf{1}_k \otimes \mathbf{z}_\mathbf{V})    \diag(\mathbf{s}_\mathbf{V})^{-1}
\end{align*}
where $\mathbf{1}_k \in \mathbb{Z}^{k}$ is a vector of 1s and $\otimes$ is the outer product.\footnote{If  the offset vectors are not integers we can expand the expression and still use integer matmuls for $\mathbf{Q}_{\mathbf{U}}\mathbf{Q}_{\mathbf{V}}$. For the cross terms we can use the identity  $ (\boldz_{\mathbf{u}} \otimes \mathbf{1}_k) \mathbf{Q}_{\mathbf{V}} = \boldz_{\mathbf{u}} \otimes (\mathbf{1}_k^\top \mathbf{Q}_{\mathbf{V}})$, and thus we can  still make use of integer matmuls for most of the FLOPs.
} Note, however, lower-precision matmuls {cannot} straightfowardly be used if the $\mathbf{U}$ is quantized at the \emph{column} level.

This second strategy which makes use of lower-precision matmuls can significantly improve inference latency and energy efficiency on supported hardware. For example, INT4 tensor core matmuls   can be up to four times faster than FP16 tensor core matmuls on the NVIDIA Ampere architecture,\footnote{\url{https://developer.nvidia.com/blog/nvidia-ampere-architecture-in-depth/}} while from a hardware-efficiency perspective, dedicated hardware for integer operations require much less area and energy usage than their floating-point counterparts \citep{jouppi2021ten,van2023fp8}.

\vspace{-2mm}
\subsection{Challenges in LLM Quantization}
\vspace{-2mm}
In LLMs, the majority of FLOPs are spent on dense matmuls of the form $\boldX \boldW$ where $\boldX \in \reals^{L \times d_{in}}$ are the input activations (for $L$ input tokens) and $\boldW \in \reals^{d_{in} \times d_{out}}$ are the model weights. For the Transformer architecture in particular this corresponds to the key, query, value projection layers, as well as the FFN layers. Given the sheer number of FLOPs in LLMs, inference efficiency can be improved significantly through lower-precision matmuls.

While there has been much work on post-training weight-only quantization for pretrained LLMs \citep[\textit{inter alia}]{frantar-gptq,dettmers2023case,lin2023awq,kim2023squeezellm,dettmers2023spqr,chee2023quip,lee2023owq,egiazarian2024extreme}, PTQ for activations remains difficult due to the presence of {outlier channels} in LLMs trained with standard precision \citep{dettmers2022llmint8,xiao2023smoothquant}. Informally, outlier channels are a set of input channels (i.e., columns of $\boldX$) whose values are many orders of magnitudes higher than the others, and have been shown to be crucial for performance \citep{kovaleva2021bert}. If one were just interested in quantizing $\boldX$ independently, outlier channels could be managed  by quantizing each column of $\boldX$ separately such that the scaling factor associated with an outlier channel is commensurate. However, as outlined in the previous section this would not enable the use of lower-precision matmuls, which requires $\boldX$ to be quantized by (at most) rows; unfortunately  row-level (i.e., per-token) quantization results in significant performance degradations \citep{xiao2023smoothquant}.

\vspace{-2mm}
\subsection{Quantization-Aware Training}
\vspace{-2mm}

\begin{figure*}[t]
    \vspace{-10mm}
    \centering
    \includegraphics[width=0.95\textwidth]{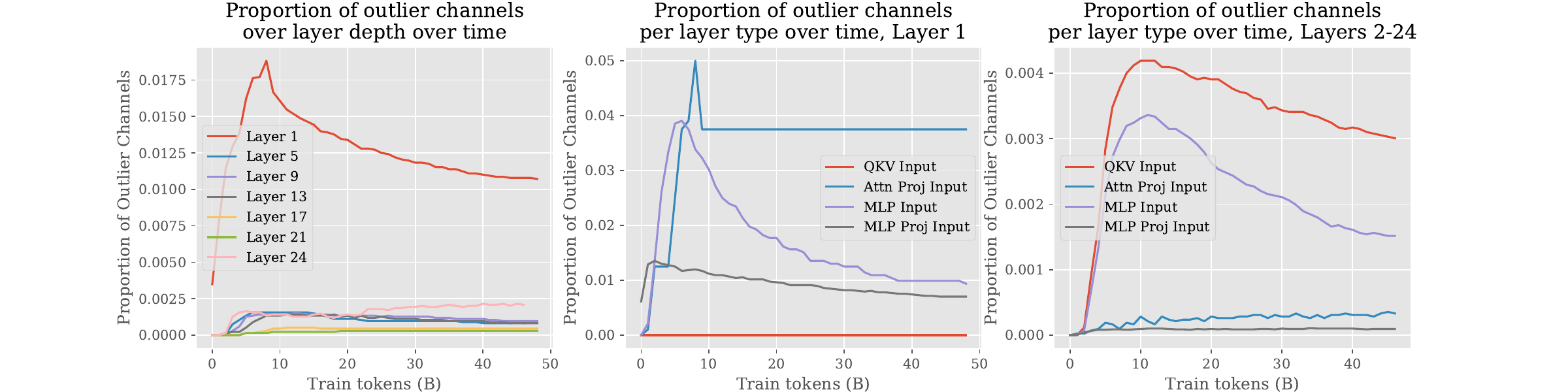}
    \vspace{-2mm}
    \caption{Frequency of outlier channels over the course of training. (Left) Proportion of outlier channels  by layer depth. Layer 1 has highest occurrence of outlier channels. (Middle) In layer 1 inputs to the attention projection layer have the most outlier channels. (Right) This is generally not the case for the other layers, where the input to the QKV project layer has the most outlier channels.}
    \label{fig:combined-outlier-model-stage}
    \vspace{-4mm}
\end{figure*}

Quantization-aware training (QAT) describes a class of techniques which aims to enable better quantization by simulating quantization during training \citep[\textit{inter alia}]{zhou2016dorefa,jacob2018quantization,Zhang_2018_ECCV,jung2019learning,jain2020trained}. While there are many methods for QAT, we use a simple modified version of PACT \citep{choi2018pact} and LSQ \citep{bhalgat2020lsq}, which learn the clip values $c^{-}$ and $c^{+}$ for the activations. This approach uses the learned clip values to perform quantization during the forward pass, and uses  the straight-through estimator   for the gradients with respect to the clip values.   While QAT has been studied extensively in the context of (typically smaller) vision models, QAT for pretraining language models with more than a billion parameters  remains less explored.

\begin{figure*}[t]
    \vspace{-15mm}
    \includegraphics[width=\textwidth]{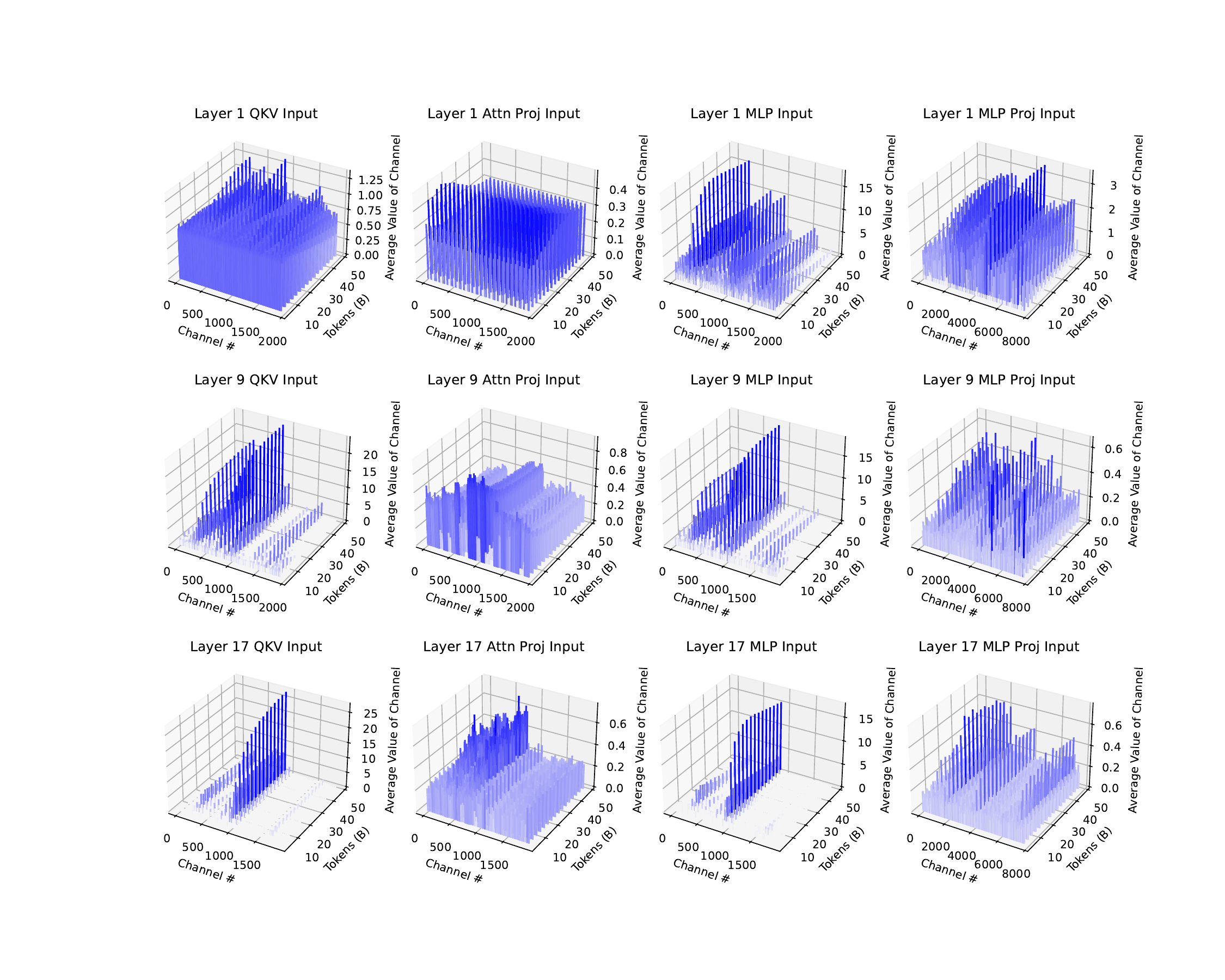}
    \vspace{-15mm}
    \caption{Trajectory of a channel's activations across 50B tokens of training. We show each channel's absolute activation value averaged across 500K tokens.}
    \label{fig:baseline-development}
    \vspace{-4mm}
\end{figure*}

\vspace{-2mm}
\section{Motivating Study: Outlier Channels in Language Models}
\vspace{-2mm}
\label{sec:analysis}
We first conduct a preliminary analysis to study the emergence of outlier channels during pretraining, with both our own and open-source models. 
For our own pretrained models, we use the standard ``pre-LayerNorm'' Transformer architecture \citep{xiong2020layer}, where given layer $l$'s input $\boldX^{(l)} \in \reals^{L \times d}$ we obtain the next layer $\boldX^{(l+1)}$ via,
\begin{align*}
    &\boldY_1 = \operatorname{LayerNorm}(\boldX^{(l)}), \hspace{4mm} \mathbf{Q},\mathbf{K},\mathbf{V} =  \boldY_1 \boldW_{QKV}, 
    \hspace{4mm} \mathbf{Y}_2 = \operatorname{softmax}(\mathbf{Q}\mathbf{K}^\top \odot \mathbf{M})\mathbf{V}, \\ 
    &\boldZ = \boldX + \mathbf{Y}_2 \mathbf{W}_O,  \hspace{4mm} 
    \mathbf{Y}_3 = \operatorname{LayerNorm}(\boldX + \mathbf{Y}_2 \mathbf{W}_O), 
    \hspace{4mm}  \mathbf{Y}_4 = \sigma(\mathbf{Y}_2\mathbf{W}_1), 
    \hspace{4mm}  \boldX^{(l+1)}= \boldZ + \boldY_4\boldW_2.
\end{align*}
Here $\boldW_{QKV} \in \reals^{d \times 3d}, \boldW_O \in \reals^{d \times d}, \boldW_1 \in \reals^{d \times 4d}, \boldW_2 \in \reals^{4d \times d}$ are learnable matrices, and the bias vectors are omitted for brevity. Our study focuses on the following activations that have been previously found to contain outlier channels:  \textit{QKV Input} ($\boldY_{1}$), \textit{Attn Proj Input} ($\boldY_{2}$), \textit{MLP Input} ($\boldY_{3}$),  \textit{MLP Proj Input} ($\boldY_{4}$). We train 1 billion parameter (24-layer model with 1920 dimensions) on 50 billion tokens from the SlimPajama dataset \citep{cerebras2023slimpajama}. We periodically collect activation statistics for all layers by running model checkpoints on (the same) 500K tokens from the C4 dataset.

First, we attempt to measure the prevalence of outlier channels aggregated by layer type and depth. For the purposes of this analysis, we name a channel an outlier if the average absolute value of the channel is over six times the average absolute value of all the input activations. This definition of an outlier channel is somewhat arbitrary, but similar definitions in the literature based on the other metrics \citep{kovaleva2021bert} generate similar results; we use this definition as opposed to definitions on the absolute values \citep{dettmers2022llmint8} to enable comparison across different layers. The results of this analysis are in \cref{fig:combined-outlier-model-stage}.
Our results generally follow what has been established in the literature: while outliers are distributed throughout depth, the layers which tend to have the most outlier channels in their input are those whose inputs are the residual stream of the network. Interestingly, we find that outlier channels emerge early in training, and rapidly become  numerous. The proportion of outlier channels within a layer then decreases gradually and eventually plateaus. 

\begin{figure*}[t]
\centering
\vspace{-14mm}
    \includegraphics[width=0.91\textwidth]{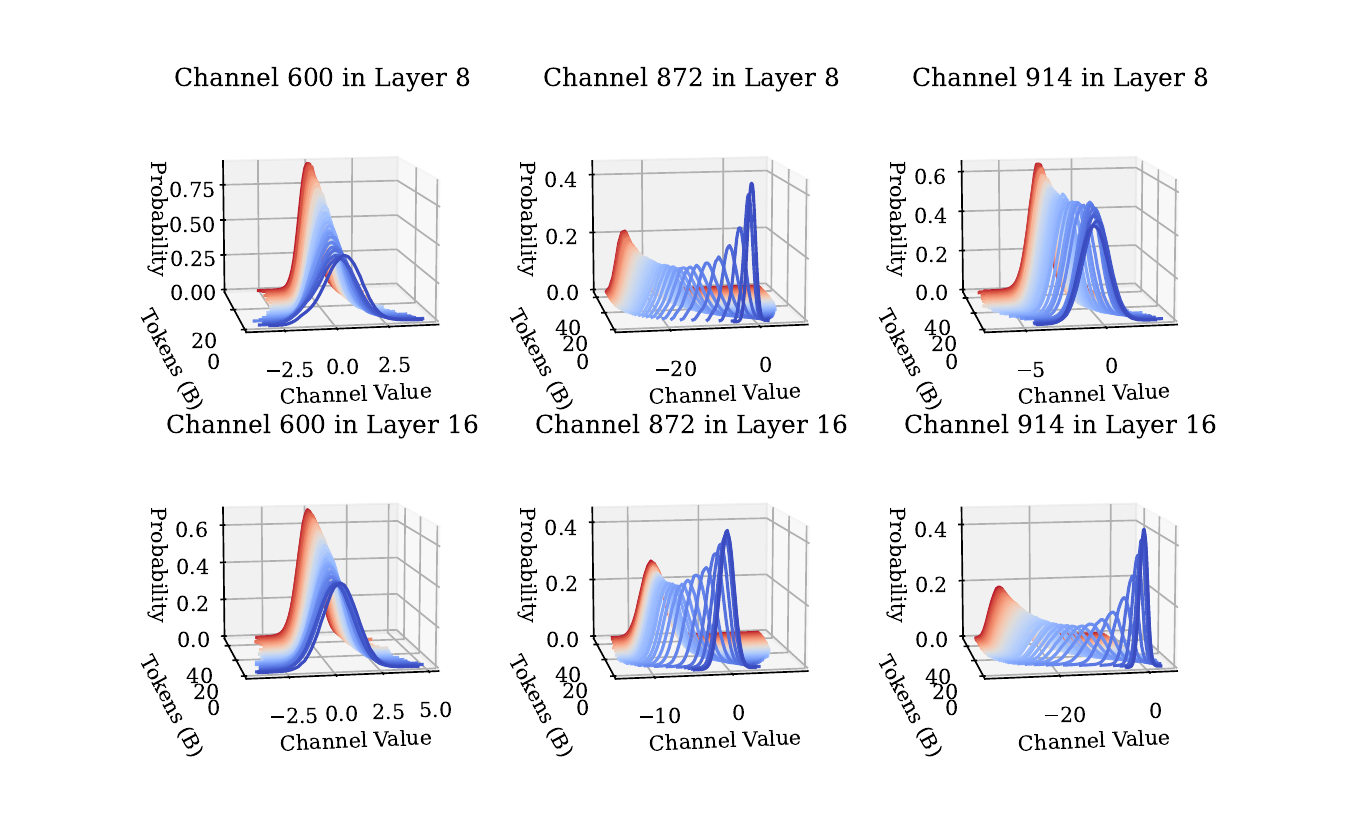}
    \label{fig:baseline-hists}
    \vspace{-8mm}
    \caption{The distribution of activations over of a non-outlier channel (left) and two outlier channels (middle, right) over training.}
    \vspace{-4mm}
    \label{fig:outlier-channel}
\end{figure*}

\begin{figure*}[!t]
    \centering
    \includegraphics[width=.93\textwidth]{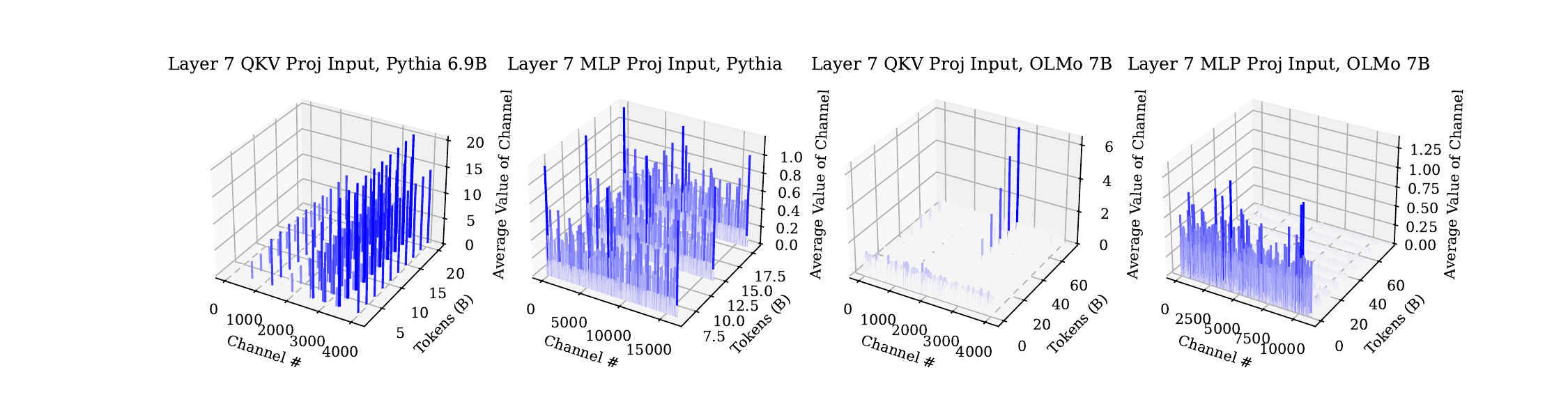}
    \vspace{-4mm}
    \caption{Activation development in two open-source models: Pythia 6.9b \citep{biderman2023pythia} and OLMo 7B \citep{groeneveld2024olmo}. We show activations for a layer that reads from the residual stream (\texttt{QKV Input}) and one that does not (\texttt{MLP Proj Input}). Note that the OLMo data includes a step-0 checkpoint (i.e., at initialization).}
    \vspace{-4mm}
    \label{7b-development}
\end{figure*}

We next perform a more granular analysis, where we analyze  the average absolute value of channels over the training of a 1B model with 50B tokens. This is shown in \cref{fig:baseline-development}. 
Within channels, we observe that the development of outliers occurs early on during training.
In most cases outliers primarily occur in layers that take as input the residual stream, although there is still significant variation in the average magnitude of channels in the input to other layers.
We take a closer  examination of the development of some the largest individual outlier channels for a particular layer in \cref{fig:outlier-channel}. Channel 600, which is not an outlier channel, has channel values that are distributed roughly as a Gaussian  with a mean of zero. The outlier channels, in comparison, have mean values that are significantly different from zero. This initial examination suggests that outlier channels are not \textit{scaled} differently than non outlier channels, but have a \textit{shifted} distribution. This potentially indicates why scaling and shifting methods, like OmniQuant \citep{shao2023omniquant}, outperform  scaling-only methods such as  SmoothQuant \citep{xiao2023smoothquant}.

\vspace{-2mm}
\paragraph{Open-source Models.} 
To validate the generality of our observations, we perform our analysis on two publicly available 7B models with public checkpoints, 
Pythia \citep{biderman2023pythia} and OLMo \citep{groeneveld2024olmo}.
In \cref{7b-development} we can see the development of activation outliers early on in the training of both models, although  the outliers in OLMo take longer to develop. Furthermore, we confirm a pattern found across the literature, that the primary place where outliers develop is not between layers in a given attention or MLP block but in the residual stream between blocks. That is, the types of layers that do or do not develop outliers are the same in both our model and the pretrained models (e.g.,  \textit{QKV Input} activations have outlier channels, while \textit{MLP Proj Input} activations do not).

\vspace{-4mm}
\section{Mitigating  Outlier Channels with Activation Regularization}
\vspace{-2mm}
Based on  insights from the previous section, we propose a simple regularization  strategy for quantizing the activations of the linear layers, where we use QAT on the input activations  and simultaneously penalize the kurtosis of the layer's outputs.
\vspace{-2mm}
\subsection{Input Activations: QAT with Learned Clip Values}
\vspace{-2mm}
As evident from \S\ref{background:quantization}, the clip values $c^{-}$ and $c^{+}$ play a key role in uniform quantization. Following PACT \citep{choi2018pact} and LSQ \citep{bhalgat2020lsq}, we treat these quantization parameters as learnable parameters and optimize them with gradient descent. Concretely, during the forward pass we run the quantization/dequantization step, as shown in \cref{alg:quantforward}. For the backward pass, we use a straight-through estimator to obtain $\nabla {\mathbf{A}}$, $\nabla c^+$, $\nabla c^-$ from $\nabla \widehat{\mathbf{A}}$ (the gradients with respect to the quantized/dequantized layer). This is shown in \cref{alg:pactback}. We will show in our experiments that quantizing during training is crucial for 4-bit quantization; just clamping the activations without quantization leads to poor performance. 

\subsection{Output Activations: Kurtosis Regularization}
\vspace{-2mm}
In our initial experiments we found that QAT on a layer's input is sufficient to train a W16A4 model that matches the performance of a W16A16. However, since we do not perform QAT for the weights, efficient deployment requires post-training weight quantization to 4 bits. While existing work has shown that weight-only PTQ to 4 bits (i.e., W16A16 $\to$ W4A16) can be done almost losslessly \citep{frantar-gptq,shao2023omniquant}, we observed this to  not be the case with QAT models, with W16A4 $\to$ W4A4 resulting in nontrivial perplexity degradations. This is due to the fact that  a model can essentially ``migrate'' the outlier channels to the corresponding rows of the weight matrix, which makes per-column weight PTQ more difficult (as shown in   \cref{fig:one}(b), bottom).

\begin{figure}[t]

 \vspace{-12mm}
    \centering
    \begin{minipage}{.47\textwidth}
        \centering
        \begin{algorithm}[H]
            \caption{QAT forward pass}
                \footnotesize
            \label{alg:quantforward}
            \begin{algorithmic}
                \REQUIRE $\mathbf{A}$, $c^-$, $c^+$, $b$, align\_zero;
                \STATE $s = \frac{2^{b}-1}{c^+ - c^-}$ 
                \IF{align\_zero} 
                    \STATE $z = \round(s \times c^-)$ 
                \ELSE 
                    \STATE $z = 0 $
                \ENDIF
                \STATE $\mathbf{Q}_{\mathbf{A}} = \round(s \times \text{ clamp }(\mathbf{A}, c^-, c^+)  + z)$
                \STATE $\widehat{\mathbf{A}} = \frac{1}{s} (\mathbf{Q}_{\mathbf{A}} - z)$
                \RETURN $\widehat{\mathbf{A}}  $
            \end{algorithmic}
        \end{algorithm}
        \vspace{-2.5mm}
        \caption{The forward and backward passes of QAT. Here $\mathbf{A}$ is the activation tensor, $b$ is the bit width, $c^-$ and $c^+$ are the learned clip values, and $\nabla \widehat{\mathbf{A}}$ is the gradient with respect to $\widehat{\mathbf{A}}$.}
    \end{minipage}%
    \hspace{2mm}
    \begin{minipage}{.47\textwidth}
    
        \centering
        \begin{algorithm}[H]
        \footnotesize
            \caption{QAT backward pass}
            \label{alg:pactback}
            \begin{algorithmic}
                \REQUIRE $\mathbf{A}$, $c^-$, $c^+$, $b$, $s$, $\nabla \widehat{\mathbf{A}}$; 
                \STATE $\mathbf{Q} = s \times (\mathbf{A} - c^-)$ 
                \STATE $\mathbf{E} = (\mathbf{{Q}} - \round(\mathbf{{Q}})) / (2^b-1)$
                \STATE $\nabla \mathbf{A}_{ij} =
                  \begin{cases}
                       0 &  \text{if } \mathbf{A}_{ij} > c^+ \text{ or } \mathbf{A}_{ij} < c^-\\
                       \nabla \widehat{\mathbf{A}}_{ij} & \text{otherwise}
                  \end{cases}$
                \STATE $\mathbf{C}^{+}_{ij} = 
                   \begin{cases}
                      \nabla \widehat{\mathbf{A}}_{ij} & \text{if } \mathbf{A}_{ij} > c^+ \\
                      -\ \mathbf{E}_{ij} \times \nabla \widehat{\mathbf{A}}_{ij} & \text{elif } \mathbf{A}_{ij} > c^- \\
                      0 & \text{otherwise}
                   \end{cases}$    
                \STATE $\mathbf{C}^{-}_{ij} = 
                   \begin{cases}
                      -\ \mathbf{E}_{ij} \times \nabla \widehat{\mathbf{A}}_{ij} & \text{if } \mathbf{A}_{ij} < c^+ \\
                      \nabla \widehat{\mathbf{A}}_{ij} & \text{elif } \mathbf{A}_{ij} < c^- \\
                      0 & \text{otherwise}
                   \end{cases}$
                \STATE $ \nabla c^+ = \sum_{ij} \mathbf{C}^+_{ij} $, \hspace{5mm} $ \nabla c^- = \sum_{ij} \mathbf{C}^-_{ij} $
                \RETURN $\nabla \mathbf{A}, \nabla c^+, \nabla c^-$
            \end{algorithmic}
        \end{algorithm}
       
    \end{minipage}
     \vspace{-4mm}
\end{figure}

One approach to mitigating these \emph{outlier weights} would be to directly regularize the weights via QAT or some other approach (e.g., $\ell_\infty$-norm regularization). However, we found these direct regularization approaches to result in much worse performance and/or unstable training. We thus adopt a more indirect regularization strategy, exploiting the fact that high input channel weights typically lead to a layer's \emph{outputs} having  outliers, i.e., the output distribution is heavy-tailed (see \cref{fig:one}). Our approach thus regularizes the output distribution's {kurtosis}, which intuitively measures how heavy-tailed a distribution is.  An  estimate of the kurtosis of a set of values $\mathbf{x} \in \reals^d$ is given by
$$
   \operatorname{Kurtosis}(\mathbf{d}) = \frac{\sum_i^k(\mathbf{x}_i-\mu)^4}{\sigma^4 + \epsilon},
$$
where $\mu$ and $\sigma$ are respectively the empirical mean and standard deviation of $\mathbf{x}$, and $\epsilon$ is a small term for numerical stability.  We multiply the sum of the kurtosis estimates for each token with  hyperparameter $\lambda$, and add the result to the cross-entropy loss. While prior work has shown the benefits of regularizing the kurtosis of a layer's activation distribution to be close to that of a uniform distribution \citep{chmiel2020robust}, regularizing the output distribution's kurtosis to make it less heavy-tailed has not been explored before to our knowledge.

\vspace{-2mm}
\subsection{Post-training Weight Quantization}
\vspace{-2mm}
After training the model to W16A4 with activation regularization on both the inputs/outputs, we experiment with two methods for quantizating the weights to 4 bits. The simplest baseline we use is round-to-nearest (RTN) quantization, which for our purposes implies per-token (for activations)\footnote{While there are more sophisticated activiation quantization approaches \citep{yuan2023rptq,chee2023quip}, these typically have additional overhead (for low-precision matmuls) and are thus not as fast as simple RTN integer quantization.} or per-output-channel (for weights) uniform min-max quantization. While the underperformance of RTN weight quantization versus more sophisticated quantization strategies that use calibration data is widely known, we deliberately include this simple data-agnostic baseline to show that activation regularization results in weights that are also easier to quantize (i.e., less perplexity degradation with RTN). Our second approach applies GPTQ \citep{frantar-gptq}, which uses a small amount of calibration data to quantize the weights, and is still near the state-of-the-art for 4-bit weight quantization.
\vspace{-4mm}
\section{Empirical Study}
\vspace{-2mm}

\begin{table*}[]\centering
\vspace{-7mm}
\footnotesize
\begin{tabular}{@{}rrrcrrrr@{}}\toprule
& \multicolumn{2}{c}{Native Activations} & \phantom{abc}& \multicolumn{4}{c}{4-bit Activations} \\
\cmidrule{2-3} \cmidrule{5-8}
Weight Precision & 16 &  4 && 4 & 4 & 3 & 3  \\ 
Weight Quantizer &None &GPTQ&&GPTQ&RTN&GPTQ&RTN \\
\midrule
\textit{Baseline}  & 23.57 & 24.10 && 113233 & 11855 & 11755 & 17187 \\ 
\textit{Activation Clamping} & 23.73 & 24.85 && 378 & 423 & 568 & 663 \\ 
\textit{Kurtosis Regularization} & 23.72 & 24.57 && 8720 & 8140 & 10235 & 19665 \\ 

\textit{QAT } & 24.30 & 25.32 && 25.32 & 27.76 & 32.56 & 46.47 \\ 
\textit{ QAT + Kurtosis Regularization} & 24.10 & 24.57 && 24.57 & 24.90 & 26.83 & 30.46 \\ 
 \midrule
\textit{Baseline} & 25.70 & 26.16 && 8430 & 10028 & 9107 & 14498 \\
\textit{Activation Clamping} & 26.38 & 27.60 && 32378 & 6852 & 26120 & 15908 \\
\textit{Kurtosis Regularization} & 26.28 & 26.95 && 7319 & 6852 & 9066 & 15908 \\
\textit{QAT} & 26.72 & 27.86 && 27.87 & 32.70 & 64.61 & 58.81 \\
\textit{QAT + Kurtosis Regularization} & 26.11 & 26.56 && 26.56 & 27.13 & 30.12 & 33.46 \\

\bottomrule
\end{tabular}
\vspace{-2mm}
\caption{Perplexity  of 1B models on C4 (top) and PTB (bottom). Native activation  are 16 bits for \textit{Baseline}, \textit{Activation Clamping},  \textit{Kurtosis Regularization}; and 4 bits for \textit{QAT}, \textit{ QAT + Kurtosis Regularization}. }
\label{tab:1B}
\vspace{-4mm}
\end{table*}

\subsection{Experimental Setup}
\vspace{-2mm}
We use the Megatron-LM \citep{shoeybi2020megatronlm} codebase  and train on the SlimPajama dataset \citep{cerebras2023slimpajama}. While the trajectory analyses in \S\ref{sec:analysis} were done for 50B tokens, due to limited compute we train for 20B tokens for these experiments. 

\vspace{-2mm}
\paragraph{Baselines.} In order to isolate the contributions of each component of our method, we compare against several baselines, on top of the standard-precision baseline. The \textit{activation clamping} baseline uses static, per-layer clipping values to clamp the input activations. To advantage this approach as much as possible, we use an ``oracle'' clipping values obtained from QAT to decide the per-layer clipping values, which was found to be more effective than grid-searching on the clipping values. In activation clamping the activations are not quantized during training, and thus this baseline isolates the effect of QAT. The \emph{kurtosis regularization} baseline applies kurtosis regularization just on the outputs, without QAT. The \emph{QAT}-only baseline just applies QAT in the input activations. 

\vspace{-2mm}
\paragraph{Hyperparameters.} All hyperparameters were tuned for our 1B W16A16 baseline and kept constant throughout experiments, except for weight decay where we selected between $\{0.1, 0.01\}$ for all methods. We use a batch size of 1M tokens, learning rate of 1.5e-4, cosine learning decay, and FP16 precision. For QAT we initialize our clipping values to $\pm 4$ for clipping value initializations, unless the layer's input is bounded. We use the same learning rate but no momentum or weight decay for clip values. For kurtosis we use 1e-5 as the regularization strength.

\vspace{-2mm}
\paragraph{Evaluation.}
We evaluate the perplexity of each model on the C4 and PTB  datasets.
We test models in three different weight quantization categories: 16 bits, 4 bits, and 3 bits. The 4-bit and 3-bit experiments test with both RTN and GPTQ.
For activations, we test in native precision (16 bits for non-QAT models, and 4 bits for the QAT models) as well as in 4 bits. For  GPTQ  we use a small amount of C4 data for calibration.

\vspace{-2mm}
\subsection{Results}
\vspace{-2mm}

We report the results of our 1B experiments on the C4 and PTB dataset in \cref{tab:1B}. 
We observe that our approach can  learn a W4A4 model that has respectable performance compared to the W16A16 baseline. We also observe that the gap between the QAT model with and without kurtosis expands as weights are quantized more and more. 
At full precision, the gap is less than 1\%. At 4 bits, this expands to between 3\% and 4\%, and at 3 bits this gap widens to 21\%.
All non-QAT methods have catastrophic performance degradations with 4-bit activations:  Activation clamping is the only method that achieves less than two orders of magnitude increase in perplexity.
In \cref{tab:downstream-eval} we perform experiments on downstream tasks for select models to validate our usage of perplexity as a proxy for downstream performance. We observe that models with similar perplexity exhibit  similar downstream performance.

We also perform a suite of experiments at the 300M scale, where we just experiment with the QAT baselines. This is shown in ~\cref{tab:300M}. We largely observe the same trends, with one exception: the gap between the QAT and QAT+Kurtosis Regularization model is  smaller than at the 1B scale.

\begin{table*}[t]\centering
\footnotesize
\vspace{-8mm}
    \begin{tabular}{rllll}
    \toprule
        Model  & Setting & HellaSwag & PIQA & ARC-easy \\ 
        \midrule
        \textit{Baseline} & {W16A16} & 32.13\% & 65.51\% & 48.32\%\\ 
        \textit{QAT} & {W16A4} & 31.79\% & 65.56\% & 47.85\%\\
        \textit{QAT + Kurtosis Regularization} & {W16A4}& 31.50\% & 64.96\% & 48.36\%  \\ 
        \bottomrule
    \end{tabular}
    \vspace{-2mm}

    \caption{Downstream evaluation of our 1B models on HellaSwag, PIQA, and ARC-easy.}
        \label{tab:downstream-eval}
    \vspace{-3mm}
\end{table*}

\begin{table*}[t!]\centering
\footnotesize
\vspace{2mm}
\begin{tabular}{@{}rrrcrrrr@{}}\toprule
& \multicolumn{2}{c}{Native Activations} & \phantom{abc}& \multicolumn{4}{c}{4b Activations} \\
\cmidrule{2-3} \cmidrule{5-8}
Weight Precision & 16 &  4 && 4 & 4 & 3 & 3  \\ 
Weight Quantizer &None &GPTQ&&GPTQ&RTN&GPTQ&RTN \\
\midrule
\textit{Baseline} & 29.23 & 30.36 && 4288 & 3864 & 4820.5 & 3923.96 \\ 
\textit{QAT} & 30.25 & 31.30 && 31.30 & 32.55 & 36.47 & 44.73 \\ 
\textit{QAT+Kurtosis Regularization} & 29.95 & 30.83 && 30.83 & 31.73 & 35.47 & 45.04 \\ 
 \midrule
\textit{Baseline}& 32.61 & 34.12 && 2974 & 2896 & 3767 & 2950 \\
\textit{QAT} & 33.56 & 34.83 && 34.83 & 34.24 & 47.51 & 51.22 \\
\textit{QAT+ Kurtosis Regularization}& 33.14 & 34.23 && 34.23 & 34.55 & 40.74 & 52.63 \\
\bottomrule
\end{tabular}
\vspace{-2mm}
\caption{Perplexity  of 300M models on C4 (top) and PTB (bottom). Native activation  are 16 bits for \textit{Baseline}, \textit{Activation Clamping},  \textit{Kurtosis Regularization}; and 4 bits for \textit{QAT}, \textit{QAT + Kurtosis Regularization}. }
\label{tab:300M}
\vspace{-2mm}
\end{table*}

\vspace{-2mm}
\subsection{Analysis}
\vspace{-2mm}

\paragraph{Post-Training Quantization of Activations.} Our method shows that QAT from scratch is effective for training a model with 4-bit activations. However, given that most available pretrained models are not trained with 4-bit activations, it would be ideal if we could take a 16-bit activation model and then finetune it with QAT to 4 bits.
To test for whether this is possible, we performed an extensive hyperparameter search for QAT finetuning on the pretrained 300M baseline model, where we finetune with QAT for 1B tokens.
Even with extensive hyperparameter tuning, QAT finetuning resulted in a W4A4  model with a 16\% degradation in perplexity over the W16A16 baseline. Upon further investigation, we found that while our QAT-pretrained models were able to learn to clip outliers without hurting performance, the QAT finetuning models struggled to do so.  Finetuning the model longer than 1 billion tokens did not improve results. 

We also tried applying OmniQuant \citep{shao2023omniquant}, a state-of-the-art weight-and-activation  method for PTQ, to go from W16A16 to W4A4. We found this approach to not perform well, with a significant degradation in perplexity with the 1B model (74.99 on C4 and 107.29 on PTB). Our degradation is larger than what has been reported for pretrained models in the original paper, which could potentially be due to our use of a smaller model (which are typically harder to quantize). Given that the outlier channels seem to emerge early in training (\S\ref{sec:analysis}), these negative results highlight the importance of early-training interventions for achieving 4-bit activation models.

\vspace{-2mm}
\paragraph{Direct Approaches for Weight Regularization.} Our use of kurtosis regularization on the output activations to mitigate the effect of ``quantization difficulty migration'' from the activations to the weights is admittedly indirect. We also experimented with more direct methods for controlling the outliers in the weights:  regularizing the kurtosis of the {weights} instead (at the tensor-level or  at the column-level); and  regularizing the weight's $l_\infty$ norm. Despite an extensive hyperparameter search, these methods led to unstable training, and we were unable to get these models to converge (unless the regularization-strength hyperparameter was so low that there was effectively no regularization). QAT on the weights also proved unsuccesful, with QAT-weight models underperforming baselines by a significant margin.

\vspace{-2mm}
\paragraph{Training Throughput.}
Our QAT approach requires modifying the forward and backward passes, which adds nontrivial overhead with an unoptimized, \texttt{torch.compile}-only implementation. This is mainly due to the reduction step in the clip val gradient in the backward pass. We thus implemented our own CUDA kernels  that perform a blockwise reduction followed by atomic additions to enable faster throughput. The throughput of our custom kernels on a single H100 node (with eight GPUs) is shown in
\cref{tab:throughput}.  We find that while there is still some reduction in throughput, it is closer to the baseline setting than the \texttt{torch.compile} implementation. Given that the numbers in \cref{tab:throughput} are from a single node, we anticipate that the actual throughput differences would be even smaller when taking into account the necessary overheads of distributed training.

\vspace{-2mm}
\section{Limitations \& Discussion}
\vspace{-2mm}

There are several limitations to our study. While we experiment with language modeling at moderate scale, we were unable to perform experiments on larger models (and train for longer) due to limited compute resources. However, we note that while the 300M parameter models did not benefit as much from the kurtosis intervention on top of QAT, at 1B there was quite a large benefit; this gives us optimism for the utility of our methods at larger scale. 

Our study targets integer quantization to 4 bits to enable the use of INT4 matmuls, which is supported by the Ampere architecture GPUs. The more recent GPU architectures (Hopper, Blackwell) unfortunately do not natively support INT4 matmuls, which limit the applicability of our approach on these GPUs. However, the latest Blackwell architecture supports FP4 computations,\footnote{\url{https://www.nvidia.com/en-us/data-center/technologies/blackwell-architecture/}} and it is possible that QAT may  improve FP4-training and moreover enable even lower-precision quantization.

 Finally, our study focuses on quantizing only the activations of inputs to linear layers, since linear matmuls consumes the majority of FLOPs during LLM inference (on moderate-length sequences). Future work could consider applying QAT to quantize the activations involved in the attention computations, which could be extremely useful in long-context settings.

\begin{table}[]
\vspace{-6mm}
\footnotesize
    \centering
    \begin{tabular}{lllll}
    \toprule
    Model size & Batch size & Baseline   & QAT (\texttt{torch.compile})  & QAT (our custom CUDA kernel) \\
    \midrule
         1B  & 1M tokens & 41913 & 20195 & 37510 \\
         3B  & 2M tokens & 15161 & 7519 & 13142 \\
    \bottomrule
    \end{tabular}
        \vspace{-2mm}
    \caption{Throughput in terms of tokens per second (TPS) on a single node with eight H100s (higher is better). The baseline achieves approximately 50\% mean FLOPs utilization (MFU), while our kernel achieves 45\%.}
    \vspace{-4mm}
    \label{tab:throughput}
\end{table}

\vspace{-2mm}
\section{Conclusion}
\vspace{-2mm}
We study outlier channels in language models from a pretraining perspective. We show that these channels emerge early in pretraining, and are moreover particularly numerous in activations with residual streams. Based on these findings, we propose a simple strategy for mitigating the effect of these outlier channels through activation regularization. We regularize the input activations with QAT plus learned clip values, and  we further regularize the output activations via the kurtosis.  Our approach is able to learn a  W4A4 language model at reasonable scale (1 billion parameters trained on 20B tokens) that is competitive with the standard-precision W16A16 baseline.

\vspace{-2mm}
\section*{Acknowledgments}
\vspace{-2mm}
This study was supported by  MIT-IBM Watson AI Lab. 

\bibliography{refs}
\bibliographystyle{colm2024_conference}

\end{document}